\documentclass[10pt,twocolumn,letterpaper]{article}
\usepackage[pagenumbers]{cvpr}
\usepackage{graphicx}
\usepackage{amsmath,amssymb}
\usepackage{booktabs}
\usepackage{xspace}
\usepackage{iftex}
\ifXeTeX
  \usepackage{fontspec}
\fi
\newcommand{\method}{RECLAIM\xspace}

\newcommand{\PASS}{\mathrm{PASS}}

\definecolor{cvprblue}{rgb}{0.21,0.49,0.74}
\usepackage[pagebackref,breaklinks,colorlinks,allcolors=cvprblue]{hyperref}
\hypersetup{pdftitle={RECLAIM: Robot-Vision-Informed Adversarial Patch Defense via Embodied Attention Feedback},pdfauthor={Jiayou Zhu, Peng Guo, Litian Zhang, Xinze Shi, Yuting Zhang, Zelong Yu, Yize Tian, Jirong Liu, Zifeng Kang},pdfsubject={Research preprint}}
\def\confName{CVPR}
\def\confYear{2027}
\title{RECLAIM: Robot-Vision-Informed Adversarial Patch Defense via Embodied Attention Feedback}
\author{%
\normalsize\begin{tabular}{c}
Jiayou Zhu$^{1}$, Peng Guo$^{1}$, Litian Zhang$^{1}$, Xinze Shi$^{1}$, Yuting Zhang$^{1}$\\
Zelong Yu$^{2}$, Yize Tian$^{1}$, Jirong Liu$^{1}$, Zifeng Kang$^{1,*}$\\[2pt]
$^{1}$Beijing University of Posts and Telecommunications\quad $^{2}$Beihang University\\[2pt]
{$^{*}$Corresponding author.}
\end{tabular}
}

\makeatletter
\patchcmd{\@maketitle}{\vskip .375in}{\vskip 0pt}{}{\errmessage{Title top-space patch failed}}
\patchcmd{\@maketitle}{\vspace*{24pt}}{\vspace*{2pt}}{}{\errmessage{Title author-space patch failed}}
\patchcmd{\@maketitle}{\vspace*{12pt}}{\vspace*{0pt}}{}{\errmessage{Title bottom-space patch failed}}
\patchcmd{\@maketitle}{\vskip .5em}{\vskip 0pt}{}{\errmessage{Title footer-space patch failed}}
\makeatother
\begin{document}
\maketitle
\begin{abstract}
Vision-language-action policies control embodied agents through a perception--action loop grounded in physical interaction, where adversarial patches can corrupt action predictions and thereby alter subsequent observations. To mitigate these threats, many existing defenses rely on policy adaptation or auxiliary models, but require additional training or inference. We introduce RECLAIM, a training-free, inference-time defense that uses a frozen VLA's own attention response to local visual interventions to guide purification during closed-loop embodied manipulation. Our key insight is that attention changes induced by local edits provide feedback for choosing where and how to purify as the robot interacts with its environment. RECLAIM combines this feedback with robot geometry to refine patch localization and select local treatments, then revalidates them as the agent's embodied actions change the observed scene. We evaluate RECLAIM against patches generated by four adversarial attack methods across multiple manipulation task suites under frozen VLA control. Under adversarial patch attacks, RECLAIM succeeds in at least one rollout on 95\% of the evaluated tasks and raises rollout success on LIBERO-Object from 0\% without defense to 70.0\%. Its mean decision time is only 548.9 ms per control step. These results highlight policy-guided visual purification as an effective approach to defending embodied agents against adversarial patches and recovering closed-loop task execution.
\end{abstract}

\section{Introduction}
\label{sec:introduction}

\begin{figure}[t]
\centering
\includegraphics[width=\columnwidth]{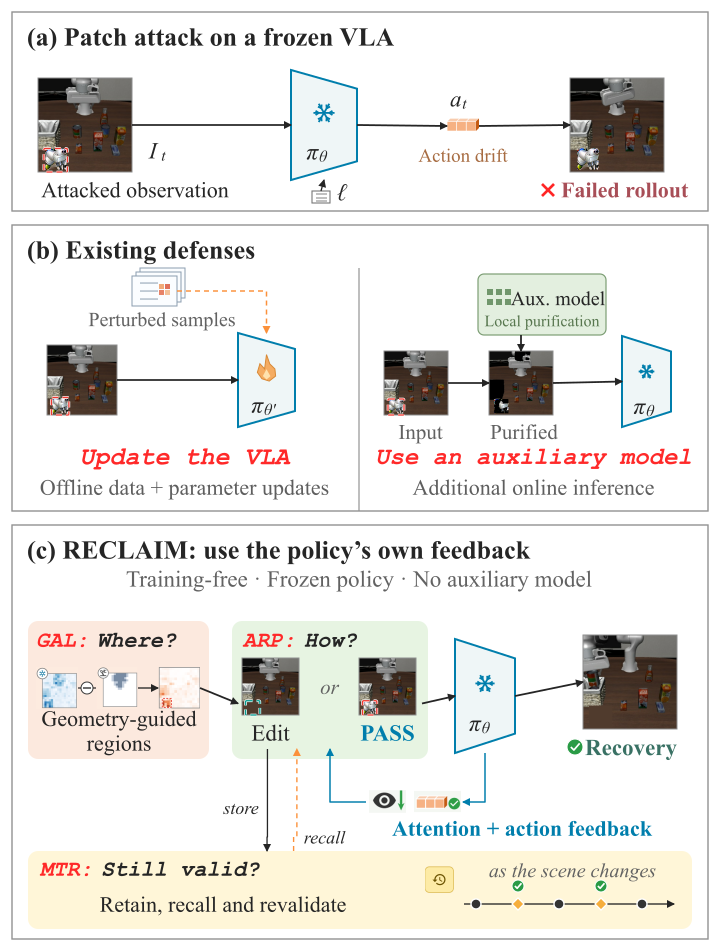}
\caption{\textbf{Policy-guided defense against adversarial patches.} (a) A patch can redirect a frozen VLA's actions and disrupt task execution. (b) Existing defenses update the VLA or invoke an auxiliary model. (c) RECLAIM uses the frozen policy's own attention and action feedback to guide localization, purification, and temporal revalidation.}
\label{fig:at_a_glance}
\end{figure}

Vision-language-action (VLA) models use visual observations and language instructions to predict robot actions~\cite{rt2,openvla}. Their visual representations determine the scene information available for action prediction, so corrupted perception can change how the robot interacts with its surroundings. The security of embodied deployment therefore depends on the integrity of visual perception. Adversarial patches introduce localized patterns to redirect visual predictions~\cite{brown2017patch}. In VLA control, they exploit this dependence by altering the camera observation: the altered pixels change the encoded visual features and can redirect action predictions without changing policy parameters~\cite{uada,edpa}. The resulting interference can propagate through the perception--action loop. An erroneous movement changes the robot and object state, altering the next observation and potentially compounding the initial disturbance. Protecting VLA deployment thus requires recovering instruction-following control as interaction continues.

Existing defenses address this vulnerability through auxiliary perception or adversarial training, introducing additional deployment requirements~\cite{pad,edpa}. PAD refines patch localization with pretrained segmentation~\cite{pad}, adding model inference to input processing. Xu et al. fine-tune the visual encoder to align clean and adversarial representations~\cite{edpa}, requiring perturbed training samples and parameter updates before deployment. Run-time intervention avoids updating the controlled policy but can retain external dependencies: BYOVLA identifies task-irrelevant regions through external scene interpretation and probes policy sensitivity before editing them~\cite{byovla}. Its scene interpretation adds auxiliary inference to the intervention process. These requirements increase online computation or offline adaptation work, complicating deployment when resources or opportunities for retraining are limited. Reducing these dependencies calls for a defense that guides localization and treatment selection using the VLA's available responses.

This requirement raises a spatial challenge of distinguishing suspicious attention from legitimate manipulation responses, and a temporal challenge of maintaining suitable interventions as the scene changes. Analyses of VLA visual attention show that patches can attract responses away from the arm and relevant objects~\cite{edpa}. Yet the gripper and manipulated object also attract attention, making response strength alone ambiguous. To interpret the robot-related part of this response, known robot geometry can account for attention associated with the arm and gripper, leaving residual concentrations for inspection. For each proposed region, the defense must then determine the most suitable treatment while retaining visual evidence needed for manipulation. As motion changes the robot--object relationship, earlier choices require renewed assessment so that useful treatments remain available and unsuitable ones can be replaced. Sustained recovery therefore depends on spatial interpretation, treatment selection, and temporal updating.

Our key insight is that the defended VLA can assess local interventions through changes in its own attention and action predictions. Comparing alternative views of the same observation with the same frozen policy makes these responses comparable in the current task context, even when a corresponding clean observation is unavailable. Repeating this assessment as interaction progresses connects purification to the policy's changing perception and action.

Figure~\ref{fig:at_a_glance} introduces RECLAIM, which couples three components for inference-time protection of the perception--action loop. \textbf{Geometry-Guided Attention Localization (GAL)} combines persistent attention with projected arm and gripper geometry to expose suspicious residual concentrations. \textbf{Attention-Response-Guided Purification (ARP)} compares local treatments through attention redistribution and action uncertainty. \textbf{Memory-Assisted Temporal Revalidation (MTR)} retains earlier treatments for comparison with fresh proposals and reassesses their suitability as the scene changes. The selected treatment modifies the observation supplied to the same frozen VLA for action generation.

We evaluate RECLAIM on frozen OpenVLA across 40 tasks in four LIBERO suites~\cite{libero} and four patch attacks~\cite{uada,edpa,rfas}. Under matched UADA conditions, RECLAIM raises rollout success from 0.0\% without defense to 52.92\%, versus accelerated PAD's 40.83\%. It also improves mask alignment and uses 77.31\% less mean decision time than accelerated PAD. With unchanged defense settings, success ranges from 38.33\% to 54.58\% across the four attacks. Ablations support the contributions of response-guided treatment selection and temporal revalidation.

Our contributions are as follows.
\begin{itemize}
    \item We formulate input-side defense for a frozen VLA as stateful local visual intervention, using the policy's attention and action responses to guide recovery of instruction-following control.
    \item We develop \method, coupling geometry-guided localization, response-guided purification, and memory-assisted temporal revalidation to counter patch interference and sustain instruction-following control as embodied interaction changes the scene.
    \item We demonstrate task recovery across four patch attacks, quantify intervention alignment and decision cost, and examine component contributions through ablations.
\end{itemize}

\section{Related Work}
\label{sec:related_work}

\paragraph{Adversarial patches in VLA control.}
Wang et al.~\cite{uada} expose the vulnerability of VLA-controlled manipulation to localized visual patches. Their Untargeted Action Discrepancy Attack (UADA) maximizes action deviations, while UPA disrupts movement directions and TMA induces targeted manipulation errors. UADA reaches a maximum average task failure rate of \(100\%\) in their simulation experiments and an attack success rate exceeding \(43\%\) in real-world trials. The resulting erratic movements show how image corruption can propagate into physical execution, threatening both task completion and safe interaction. These risks underscore the importance of defending the perception--action loop for reliable embodied deployment. This attack framework underlies our primary evaluation. Complementing these attacks on action outputs, Xu et al. introduce the Embedding Disruption Patch Attack (EDPA), which generates patches by disrupting visual--textual alignment and separating perturbed visual representations from clean ones~\cite{edpa}. They counter this representation-level interference by adversarially fine-tuning the visual encoder, requiring clean--perturbed training pairs and parameter updates. Lu et al. develop UPA-RFAS, combining feature disruption, attention dominance, and image--text misalignment to improve universal patch transfer~\cite{rfas}. We denote this attack as RFAS to distinguish it from Wang et al.'s UPA~\cite{uada}.

\paragraph{Input-side adversarial patch purification.}
Patch purification methods locate and remove corrupted image content before prediction. PAD uses semantic independence and spatial heterogeneity for localization, followed by pretrained segmentation to refine removal masks~\cite{pad}. DIFFender couples localization and restoration through text-guided diffusion with few-shot prompt tuning~\cite{diffender}, while APEX combines region extraction, blur and boundary cues, and structure-guided inpainting~\cite{apex}. These methods evaluate purification through visual recognition performance and patch-localization quality. Such criteria do not determine whether a treatment preserves the visual cues needed for a robot's next action or remains appropriate as manipulation changes the scene. Their auxiliary processing also introduces deployment dependencies: PAD invokes a segmentation model, while DIFFender requires diffusion inference and prompt adaptation. Building on local patch purification, \method addresses these control and deployment requirements by using the defended VLA itself to assess treatments in the current task context, without auxiliary perception or a separately trained defense model.

\paragraph{Policy feedback for run-time intervention.}
BYOVLA uses action sensitivity to guide observation editing without policy fine-tuning~\cite{byovla}. External VLM and segmentation models identify task-irrelevant regions; regional perturbations measure their influence on predicted actions and guide inpainting or background recoloring. \method adopts the principle of consulting the controlled policy, but extends the feedback beyond action sensitivity. Sensitivity to a region does not alone establish which purification to prefer. \method compares treatments through regional attention reduction and action uncertainty, requiring cross-operator attention agreement outside the edited region. It therefore assesses both suppression within the candidate region and consistency over the remaining scene, enabling purification selection without first assigning the suspected patch an external task-irrelevance label.

\paragraph{Spatial information in attention.}
Attention studies provide evidence that frozen vision-language models retain spatial information useful beyond image-level prediction. Kang et al.~\cite{localizationheads} identify a small set of heads whose responses localize text-referred objects. ResCLIP finds that intermediate-layer query--key attention captures spatial correspondences between image regions; incorporating these responses into the final layer improves localization for dense visual prediction~\cite{resclip}. Together, these findings support using native attention as a source of spatial cues for localization. In manipulation, strong responses can also arise from the arm, gripper, and task objects. Wang et al.~\cite{uada} suggest robot-joint relationships as a potential defense cue. Robot geometry can therefore help interpret attended regions, while comparing attention before and after local edits provides feedback about their effect on the policy's visual processing. These complementary insights motivate grounding patch purification in embodied policy feedback, connecting spatial localization and treatment selection to continued task execution.

\begin{figure*}[t]
\centering
\includegraphics[width=\textwidth]{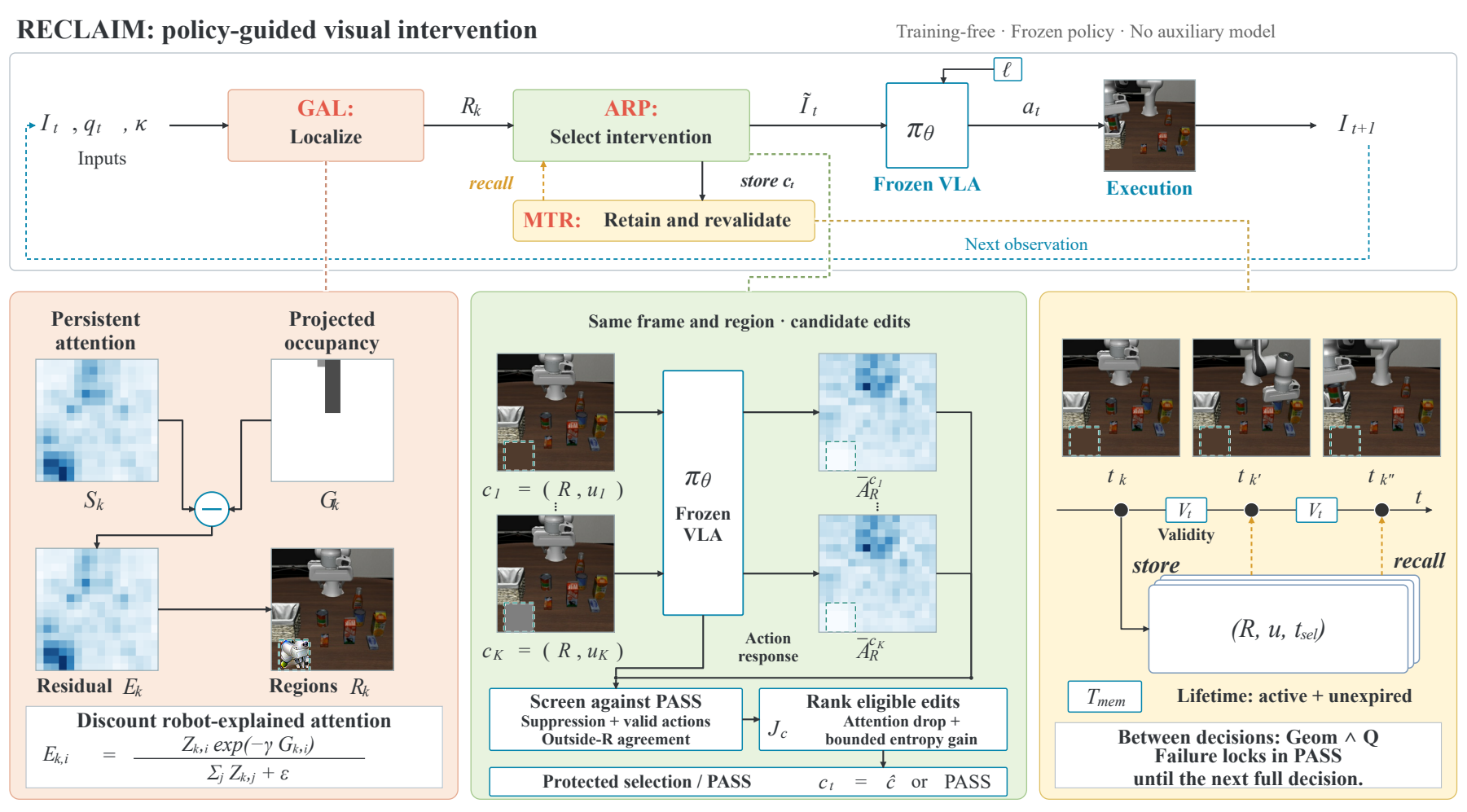}
\caption{\textbf{RECLAIM framework.} GAL combines persistent attention with projected robot occupancy to propose regions. ARP evaluates fresh edits and recalled pairs on the current image, requiring cross-family support and ranking eligible treatments by attention reduction and bounded entropy gain. Protected selection yields a local edit or $\PASS$ for action generation by the frozen VLA. MTR stores selected pairs, recalls them at full decisions, and checks geometry and action validity between decisions.}
\label{fig:overview}
\end{figure*}
\section{RECLAIM}
\label{sec:method}

\subsection{Problem Formulation}
For deployment on unfamiliar patches and changing tasks, RECLAIM operates without patch masks, paired clean images, or clean reference actions. It uses attention and action-token probabilities from a frozen vision-language-action (VLA) policy $\pi_\theta$ to select visual interventions. At step $t$, it receives a potentially patch-corrupted RGB image $I_t\in\mathbb R^{H\times W\times3}$, instruction $\ell$, robot joint configuration $q_t$, and trusted parameters $\kappa$ specifying robot geometry and camera projection. Normal images supply calibration scales before deployment.

A treatment $c=(R,u)$ pairs a region $R$ with a local purification operator $u$, including its strength. The binary mask $B_R$ specifies its spatial extent. A stateful rule $\mathcal D_{\pi_\theta}$ selects $c_t=(R_t,u_t)$ and updates state $\xi_t$, comprising attention statistics, the active treatment, and intervention memory. The resulting controller is
\begin{equation}
\begin{aligned}
(c_t,\xi_t)&=\mathcal D_{\pi_\theta}(I_t,\ell,q_t;\kappa,\xi_{t-1}),\\
\widetilde I_t&=T_{c_t}(I_t)=(1-B_{R_t})\odot I_t\\
&\qquad\qquad\quad+B_{R_t}\odot u_t(I_t),\\
a_t&=\pi_\theta(\widetilde I_t,\ell),\qquad\theta\ \text{fixed}.
\end{aligned}
\label{eq:problem}
\end{equation}
Here $\widetilde I_t$ generates the executed action $a_t$. Treatment selection includes the identity option $\PASS$, with $T_{\PASS}(I_t)=I_t$, which retains the incoming view when candidates lack support or fail protection. RECLAIM thus formulates patch defense as stateful visual intervention for instruction-following control, guided by frozen-policy responses.

\subsection{Overview}
RECLAIM links three components through attention-guided intervention (Fig.~\ref{fig:overview}). \emph{Geometry-Guided Attention Localization (GAL)} combines persistent attention with robot geometry to propose regions for inspection. \emph{Attention-Response-Guided Purification (ARP)} tests local edits and uses attention redistribution and action uncertainty to select purification. \emph{Memory-Assisted Temporal Revalidation (MTR)} reevaluates selected region--operator pairs as manipulation changes the scene, allowing useful earlier alternatives to compete with new proposals.

At full decisions $t_k=kn$, where $k$ indexes decisions and $n$ is their control-step interval, fresh and recalled treatments are evaluated on the current image. The selected view supplies the action in Eq.~\ref{eq:problem}, and subsequent observations support checks and reselection. This connects region proposals and intervention responses to control over time.

\subsection{Geometry-Guided Attention Localization}
GAL combines temporal persistence and robot geometry to prioritize sustained attention after discounting responses explained by the arm and gripper.

We use the classification-token and image-patch structure of the visual transformer~\cite{vit}, aligning its query/key responses with observation coordinates, obtaining map $L_k$. Minimum subtraction $\widehat L_k=L_k-\min_iL_{k,i}$ removes a common offset, and sum normalization $g_k=\widehat L_k/(\sum_i\widehat L_{k,i}+\epsilon)$ makes response scales comparable across decisions. Here $i$ indexes grid cells and $\epsilon>0$ stabilizes division. Moving mean $\mu_k$ and second moment $v_k$ accumulate persistence as $\mu_k=(1-\alpha)\mu_{k-1}+\alpha g_k$ and $v_k=(1-\alpha)v_{k-1}+\alpha g_k^2$, where $\alpha$ controls adaptation to new observations. The standard deviation $\sigma_k=\sqrt{\max(v_k-\mu_k^2,0)}$ gives stability response $S_{k,i}=\mu_{k,i}/(\sigma_{k,i}+\epsilon)$, prioritizing sustained attention relative to temporal fluctuation. Moments initialize from each episode's first observation.

Using $(q_{t_k},\kappa)$, we project arm and gripper core and surrounding guard masks $C_{k,h}$ onto this grid, with $h$ indexing masks. Soft occupancy $G_k=\operatorname{clip}(\sum_h\omega_hC_{k,h},0,1)$ weights their contributions by $\omega_h$ and bounds overlap values within $[0,1]$. The nonnegative response $Z_{k,i}=S_{k,i}-\min_jS_{k,j}$ removes the stability map's common offset before geometric discounting:
\begin{equation}
E_{k,i}=\frac{Z_{k,i}\exp(-\gamma G_{k,i})}
{\sum_j Z_{k,j}+\epsilon}.
\label{eq:residual}
\end{equation}
Here $j$ indexes grid cells and $\gamma$ controls occupancy discounting. The pre-discount denominator preserves reduced robot-explained attention in $E_k$ for region selection. Thresholding positive residuals yields connected-component boxes, which are screened by area, response mass, and robot overlap. Ranking by residual mass retains at most $M$ regions in $\mathcal R_k$ for the intervention comparisons in ARP.

\subsection{Attention-Response-Guided Purification}
Given regions $\mathcal R_k$ proposed by GAL, ARP selects purification by comparing local edits. Reducing regional attention dominance aims to release patch-captured attention, while concentrated action predictions and consistent attention elsewhere favor continued task control. An operator set $\mathcal U$ forms fresh treatments $\mathcal C_{\mathrm{new},k}=\mathcal R_k\times\mathcal U$. Each $c=(R,u)$ generates $T_c(I_t)$ for comparison with the untreated image and recalled treatments from MTR.

\paragraph{Measuring intervention responses.}
For each view, we aggregate classification-token attention to image patches and normalize the spatial weights to sum to one, obtaining distribution $A^c$. This enables regional comparisons across edits. The untreated distribution is $A^0=A^{\PASS}$. Let $w_{R,i}$ be the fraction of patch $i$ covered by $R$ in policy-input coordinates. Then $m_c=\sum_iw_{R,i}A_i^c$ measures attention assigned to the edited region, with baseline $m_0(R)=\sum_iw_{R,i}A_i^0$. Screening requires baseline mass $m_0(R)\geq\tau_m$ and relative reduction $r_c\geq\tau_r$ (Eq.~\ref{eq:score}), focusing evaluation on edits of attended regions. Action-quality condition $Q_t(c)$ further requires valid action tokens and sufficient probability mass on the action vocabulary, so eligible edits support a usable action prediction.

\paragraph{Establishing cross-operator support.}
Useful objects also attract attention, motivating a check of how distinct treatments redistribute attention over the surrounding scene. Let $\mathcal W_t(c)$ contain screened treatments $d=(R,u')$ from a different operator family that produce an image distinct from $c$'s. Define sum normalization as $\mathcal N(x)=x/\sum_i x_i$. Removing the regional weights gives $\bar A_R^c=\mathcal N((1-w_R)\odot A^c)$, the outside-region attention distribution under $c$; $\bar A_R^d$ is computed identically under supporter $d$. This tests whether different edits leave the policy attending similarly to the surrounding scene. A screened candidate enters eligible set $\mathcal E_t$ when
\begin{equation}
\exists d\in\mathcal W_t(c):\quad
\operatorname{JS}(\bar A_R^c,\bar A_R^d)\leq\tau_{\mathrm{JS}}.
\label{eq:consensus}
\end{equation}
Jensen-Shannon divergence~\cite{lin1991divergence} quantifies agreement between these outside-region attention distributions. Normal-image calibration sets $\tau_{\mathrm{JS}}$.

\paragraph{Ranking and selecting a treatment.}
Among eligible edits, regional attention reduction measures the release of attention dominance, while lower action entropy favors a concentrated next-action prediction. For treatment $c$, let $p_j^c(b)$ be the probability of action token $b$ at position $j$, conditioned on that view's decoded prefix and renormalized over action vocabulary $\mathcal V_a$. Across $L_a$ action positions, normalized Shannon entropy~\cite{shannon} is $H_c=-\sum_{j=1}^{L_a}\sum_{b\in\mathcal V_a}p_j^c(b)\log p_j^c(b)/(L_a\log|\mathcal V_a|)$. Relative to untreated entropy $H_{\PASS}$, we define
\begin{equation}
\begin{aligned}
r_c&=\frac{m_0(R)-m_c}{\max(m_0(R),\epsilon_m)},\\
\Delta H_c&=H_{\PASS}-H_c,\\
J_c&=r_c+\lambda_H\frac{\Delta H_c}{s_H+|\Delta H_c|}.
\end{aligned}
\label{eq:score}
\end{equation}
Here $r_c$ rewards attention reduction and positive $\Delta H_c$ rewards reduced action uncertainty. Normal-image calibration sets entropy scale $s_H$ to account for ordinary variation, while saturation bounds its contribution by $\lambda_H$. Therefore, attention reduction remains central to $J_c$, with uncertainty refining its ranking. The constant $\epsilon_m>0$ stabilizes division.

For nonempty $\mathcal E_t$, the winner $\hat c=\arg\max_{c\in\mathcal E_t}J_c$ undergoes protection $P_t$ before adoption. It combines stronger attention support for disruptive fresh edits with action-deviation checks for recalled hard-gray treatments, relating disruption to response support. The resulting selection is
\begin{equation}
\operatorname{Select}_t(\mathcal C)=
\begin{cases}
\hat c,&\mathcal E_t\ne\emptyset\ \land\ P_t(\hat c),\\
\PASS,&\text{otherwise}.
\end{cases}
\label{eq:selection}
\end{equation}
Here $\mathcal C$ supplies candidates for eligibility and ranking. The output determines the view generating $a_t$ and the region--operator pair passed to temporal revalidation.

\subsection{Memory-Assisted Temporal Revalidation}
\label{sec:temporal}
As motion changes the scene addressed by ARP, MTR balances treatment continuity with renewed evaluation. It maintains an active pair and a bank $\mathcal M_{k-1}$ of recently selected pairs. At $t_k$, recall collects the active pair and unexpired bank entries that satisfy current robot-overlap constraints. Removing duplicates against fresh proposals forms $\mathcal H_k$, allowing previously selected regions to reenter comparison alongside current localization:
\begin{equation}
c_{t_k}=\operatorname{Select}_{t_k}
\bigl(\mathcal C_{\mathrm{new},k}\cup\mathcal H_k\bigr).
\label{eq:revalidation}
\end{equation}
ARP rescores each recalled pair on the current image. Its stored operator is an execution candidate, with additional operators providing cross-family support for Eq.~\ref{eq:consensus}. Thus current responses determine its continued suitability.

The bank stores up to $B$ distinct regions as $(R,u,t_{\mathrm{sel}})$, with latest-selection time $t_{\mathrm{sel}}$. Lifetime $T_{\mathrm{mem}}$ retains entries satisfying $t_k-t_{\mathrm{sel}}\leq T_{\mathrm{mem}}$. Purification updates the selected pair and timestamp, retaining the most recently selected regions. A $\PASS$ decision expires stale entries. The bank starts empty, and the active treatment starts as $\PASS$.

Between full decisions, $V_t(c)=\mathsf{Geom}_t(c)\land Q_t(c)$ combines current robot-projection compatibility with ARP's action-quality condition. To sustain treatment through scene changes, valid treatments continue as $c_t=c_{t-1}$, while failure sets $c_t=\PASS$ for the interval's remainder. Memory supports reselection at the next full decision. Together, localization, purification, and temporal revalidation adapt the input for the executed action in Eq.~\ref{eq:problem}.

\twocolumn[{\begin{minipage}{\textwidth}
\centering
\captionof{table}{\textbf{Main results on frozen OpenVLA policies.} SR measures rollout success; Tasks is the share of tasks with at least one success. Mask IoU averages over executed control steps. ARR normalizes adversarial recovery by the matched clean--attack performance gap. Dashes mark inapplicable metrics. Methods share initial states and patch placements within each input condition.}
\label{tab:main}
\small
\setlength{\tabcolsep}{3pt}
\begin{tabular*}{\textwidth}{@{\extracolsep{\fill}}llrrrrrrrr@{}}
\toprule
Input & Defense & \multicolumn{4}{c}{Suite SR (\%)} & \multicolumn{4}{c}{Aggregate metrics (\%)} \\
\cmidrule(lr){3-6}\cmidrule(lr){7-10}
 & & Spatial & Object & Goal & Long & SR & Tasks & IoU & ARR \\
\midrule
\multicolumn{10}{l}{\textit{Clean observations and non-optimized patch controls}} \\
\midrule
Clean & No defense & 85.00 & 80.00 & 60.00 & 50.00 & 68.75 & 85.00 & --- & --- \\
 & RECLAIM & 85.00 & 85.00 & 70.00 & 30.00 & 67.50 & 85.00 & --- & --- \\
Shuffle & No defense & 70.00 & 83.33 & 81.67 & 43.33 & 69.58 & 97.50 & 0.00 & --- \\
 & RECLAIM & 63.33 & 86.67 & 76.67 & 18.33 & 61.25 & 92.50 & 1.54 & --- \\
Grey & No defense & 80.00 & 76.67 & 61.67 & 53.33 & 67.92 & 100.00 & 0.00 & --- \\
 & RECLAIM & 58.33 & 73.33 & 76.67 & 28.33 & 59.17 & 95.00 & 0.84 & --- \\
\midrule
\multicolumn{10}{l}{\textit{Primary adversarial-patch evaluation}} \\
\midrule
UADA & No defense & 0.00 & 0.00 & 0.00 & 0.00 & 0.00 & 0.00 & 0.00 & --- \\
 & RECLAIM, Grey-only & 30.00 & 48.33 & 53.33 & 18.33 & 37.50 & 85.00 & 32.04 & 54.55 \\
 & PAD (batched MI) & 26.67 & 61.67 & 56.67 & 18.33 & 40.83 & 80.00 & 37.69 & 59.39 \\
 & APEX (O/G) & --- & 41.67 & 53.33 & --- & 47.50 & 100.00 & 30.48 & 67.86 \\
 & RECLAIM, Telea-only & \textbf{38.33} & \textbf{71.67} & 46.67 & 35.00 & 47.92 & 92.50 & 49.36 & 69.70 \\
 & RECLAIM & \textbf{38.33} & 70.00 & \textbf{65.00} & \textbf{38.33} & \textbf{52.92} & \textbf{95.00} & \textbf{50.83} & \textbf{76.97} \\
\bottomrule
\end{tabular*}
\end{minipage}
\vspace{12pt}}]
\section{Experiments}
\label{sec:experiments}

We evaluate task recovery, computational efficiency, component contributions, and generalization. Table~\ref{tab:main} summarizes how RECLAIM's policy-guided interventions affect execution on clean observations, non-optimized patch controls, and the primary UADA attack.

\subsection{Experimental Setup}
\label{sec:experimental_setup}
\paragraph{Tasks and placement.}
We evaluate frozen OpenVLA policies~\cite{openvla} on the four LIBERO suites~\cite{libero}, covering ten tasks per suite to test defense across scene layouts, object interactions, goals, and long-horizon manipulation. Different fixed initial states per task and three patch placements per state vary the starting configuration and spatial interference while preserving matched conditions across methods. This gives 80 clean rollouts and 240 rollouts per patch condition. OpenVLA fuses DINOv2~\cite{dinov2} and SigLIP~\cite{siglip} visual features~\cite{openvla}; their pretrained weights remain frozen during defense. Control horizons are 220, 280, 300, and 520 steps for Spatial, Object, Goal, and Long, with early termination on success.

To vary spatial interference under these matched task conditions, we follow UADA's initial-visibility placement principle~\cite{uada}. Seeded sampling (20261116) distributes $50\times50$ patch footprints across spatially separated admissible background regions, relocating infeasible slots under the initial-view constraints. Positions are frozen before evaluation and remain fixed in image coordinates.

\paragraph{Comparison methods and input controls.}
PAD~\cite{pad} combines MI/compression heatmaps, SAM ViT-L~\cite{sam} refinement, and black-mask removal. NumPy~\cite{harris2020array} sliding-window views and histogram reductions accelerate MI while preserving heatmaps and mask selection; all PAD results use this implementation. APEX~\cite{apex} uses SAM-based region extraction and LaMa inpainting.

The shuffle control applies one fixed joint RGB-pixel permutation to the UADA texture (seed 202). Grey replaces it with RGB $(127,127,127)$. Both preserve UADA's patch area and placements, testing intervention effects on non-optimized visual perturbations alongside clean inputs.

\paragraph{RECLAIM configuration.}
We set $n=20$, $M=5$, $\alpha=0.2$, $\gamma=1$, $\lambda_H=0.3$, $B=3$, and $T_{\mathrm{mem}}=120$. The nine operators are gray blending at strengths 0.5 and 0.8, hard gray replacement, Gaussian-blur blending at strengths 0.6 and 1.0, global-mean blending at strength 0.8, neighborhood-median filling, median filtering, and Telea inpainting~\cite{telea}. Blur uses a $7\times7$ kernel with standard deviation 2, median filtering a $7\times7$ kernel, and Telea radius 3. Telea-only provides a representative inpainting comparison by restricting the executable operator while retaining the remaining ARP and MTR rules.

\paragraph{Evaluation metrics.}
Rollout success rate (SR) is the percentage of evaluated rollouts that complete the task. Task coverage (Tasks) is the percentage of evaluated tasks with at least one successful rollout. Mask IoU measures the intersection of the purification and patch regions relative to their union, averaged over executed controls. SR differences use percentage points (pp). Attack recovery is normalized to the matched clean--attack performance gap:
\begin{equation}
\mathrm{ARR}=\frac{s_{\mathrm{def,atk}}-s_{\mathrm{none,atk}}}
{s_{\mathrm{none,clean}}-s_{\mathrm{none,atk}}},
\label{eq:arr}
\end{equation}
where $s$ denotes SR and the denominator is positive. Clean outcomes are matched by task and initial state and reused across the associated patch placements.

\subsection{RQ1: Recovery of Task Execution}
\label{sec:recovery}
\paragraph{Recovery under UADA.}
RECLAIM raises SR from 0.00\% to 52.92\%, exceeding accelerated PAD by 12.08 pp and achieving higher SR in every suite. Task coverage improves from PAD's 80.00\% to \textbf{95.00\%}. RECLAIM's 76.97\% ARR exceeds PAD's 59.39\%, recovering a larger share of the attack-induced performance gap. Its executed masks achieve higher IoU (50.83\% versus 37.69\%) while modifying less of the image (6.87\% versus 10.96\%), indicating more concentrated interventions along the respective trajectories. RECLAIM also exceeds APEX by 28.33 pp on Object (70.00\% versus 41.67\%) and 11.67 pp on Goal (65.00\% versus 53.33\%). Together, these comparisons support the advantage of policy-guided intervention over conventional patch purification for robotic task recovery.

The Grey-only control reaches 37.50\% SR and 85.00\% task coverage, showing the additional recovery obtained by combining localization with treatment selection and temporal revalidation. Telea-only reaches 47.92\% SR and 92.50\% task coverage. Even this effective single-operator treatment leaves a 5.00 pp SR gap to full RECLAIM. Selecting among alternative operators allows ARP to address conditions that committing to Telea alone does not resolve.

\paragraph{Clean and non-optimized inputs.}
Across Spatial, Object, and Goal, clean SR increases from 75.00\% to 80.00\%, while shuffle and grey decrease by 2.78 and 3.33 pp. RECLAIM therefore preserves aggregate clean task execution on these suites, with smaller changes under non-optimized perturbations. Long contributes most of the overall decrease: clean SR changes from 50.00\% to 30.00\%, and shuffle and grey each decrease by 25.00 pp. Across all four suites, clean task coverage remains 85.00\%, with overall SR changes of $-1.25$, $-8.33$, and $-8.75$ pp for clean, shuffle, and grey. The principal intervention cost is concentrated in Long; aggregate clean execution is preserved in the other three suites. Long's lower clean SR suggests greater baseline difficulty; its longer action sequences may also amplify small perception or purification errors, making accurate intervention harder to sustain throughout execution.

\subsection{RQ2: Computational Efficiency}
\label{sec:efficiency}
RECLAIM averages 548.90\,ms per control ($2.07\times$ the undefended 265.17\,ms), versus accelerated PAD's 2,419.64\,ms ($9.12\times$). This is \textbf{77.31\% less mean decision time}, a $4.41\times$ ratio. APEX averages 6,264.26\,ms on Object/Goal (Table~\ref{tab:runtime}).

\begin{center}\begin{minipage}{\columnwidth}
\centering
\captionof{table}{\textbf{UADA inference cost on RTX 4090 GPUs.} Mean decision time per control and separate policy/auxiliary-process peak allocated memory. PAD uses CPU-batched MI.}
\label{tab:runtime}
\small
\setlength{\tabcolsep}{3pt}
\begin{tabular}{lrrr}
\toprule
Defense & Latency (ms) & Policy (GiB) & Aux. (GiB) \\
\midrule
No defense & 265.17 & 14.42 & 0.00 \\
PAD & 2,419.64 & 14.42 & 4.30 \\
APEX (O/G) & 6,264.26 & 14.42 & 4.49 \\
RECLAIM & 548.90 & 14.43 & 0.00 \\
\bottomrule
\end{tabular}
\end{minipage}\end{center}

RECLAIM averages 262.94\,ms on ordinary controls and 5,853.74\,ms at full decisions, amortizing candidate evaluation. PAD and APEX require auxiliary-model processes peaking at 4.30 and 4.49\,GiB allocated, respectively (PAD: 4.71\,GiB reserved). RECLAIM requires none.

\paragraph{PAD acceleration.}
CPU batching reduces first-frame MI time from 8.278 to 0.076\,s with identical outputs. Substituting original MI while fixing other costs estimates original PAD at 10.62\,s per control ($19.35\times$ RECLAIM). Reported PAD comparisons use the accelerated implementation.

\subsection{RQ3: Ablations and Component Analysis}
\label{sec:ablations}
All variants share policy checkpoints, UADA inputs, initial states, placements, and horizons; unlisted settings remain fixed in targeted ablations.

\noindent\begin{minipage}{\columnwidth}
\centering
\captionsetup{skip=2pt}
\captionof{table}{\textbf{GAL ablation (\%).} $\Delta$: ablation minus full (pp). Coverage/IoU use executed controls; PAM drop is relative patch-attention reduction at purification decisions.}
\label{tab:geometry_ablation}
\small
\renewcommand{\arraystretch}{0.85}
\setlength{\tabcolsep}{3pt}
\begin{tabular}{lrrr}
\toprule
Metric & Full & $\gamma=0$ & $\Delta$ \\
\midrule
Overall SR & 52.92 & 50.42 & $-2.50$ \\
Goal SR & 65.00 & 53.33 & $-11.67$ \\
Patch coverage & 78.76 & 68.71 & $-10.05$ \\
Mask IoU & 50.83 & 46.27 & $-4.57$ \\
PAM drop & 80.85 & 73.26 & $-7.59$ \\
\bottomrule
\end{tabular}
\end{minipage}\par\smallskip

\paragraph{Geometry-guided attention localization.}
Without geometric discounting ($\gamma=0$), robot-explained attention competes for proposals; geometric admission, ARP, and MTR remain unchanged. Reduced coverage and IoU indicate poorer patch targeting, and weaker patch-attention suppression suggests less effective purification. The accompanying 11.67 pp Goal and 2.50 pp overall SR losses connect this localization degradation through weaker purification to poorer closed-loop recovery (Table~\ref{tab:geometry_ablation}).

\vspace{-0.5\baselineskip}
\paragraph{Attention-response-guided purification.}
Grey-only replaces multi-operator selection with fixed gray blending (0.8) and the original temporal gate. Telea-only restricts execution to Telea/$\PASS$, retaining other operators as consensus witnesses. No joint scoring fills GAL's region union with RGB (127,127,127), bypassing scoring.

\noindent\begin{minipage}{\columnwidth}
\centering
\captionsetup{skip=2pt}
\captionof{table}{\textbf{ARP/MTR ablations.} SR/IoU (\%); $\Delta$: SR change from full (pp). Q/ctrl: policy queries/control (MTR comparison).}
\label{tab:ablation}
\small
\renewcommand{\arraystretch}{0.85}
\setlength{\tabcolsep}{3pt}
\begin{tabular}{llrrrr}
\toprule
Module & Configuration & SR & IoU & $\Delta$ & Q/ctrl \\
\midrule
All & Full RECLAIM & 52.92 & 50.83 & --- & 2.107 \\
\midrule
ARP & Grey-only & 37.50 & 32.04 & $-15.42$ & --- \\
 & Telea-only & 47.92 & 49.36 & $-5.00$ & --- \\
 & No joint scoring & 7.08 & 14.94 & $-45.83$ & --- \\
\midrule
MTR & No MTR & 4.17 & 16.90 & $-48.75$ & 9.572 \\
\bottomrule
\end{tabular}
\end{minipage}\par

Grey-only and Telea-only reach 37.50\% and 47.92\% SR, below full RECLAIM's 52.92\%, supporting adaptive operator choice. Direct proposal-union purification yields 7.08\% SR and 14.94\% IoU, versus full RECLAIM's 50.83\% IoU. These results support joint scoring of region--operator pairs for better-aligned region selection and higher task completion (Table~\ref{tab:ablation}).

\vspace{-0.5\baselineskip}
\paragraph{Memory-assisted temporal revalidation.}
Without MTR, GAL/ARP rerun every control without 20-step persistence or recall. SR/IoU fall to 4.17\%/16.90\%, linking treatment continuity to sustained targeting and recovery. Queries rise from 2.107 to 9.572 per control ($4.54\times$), reflecting repeated candidate evaluation avoided by temporal reuse (Table~\ref{tab:ablation}).

\raggedbottom
\subsection{RQ4: Generalization across Attacks and VLAs}
\label{sec:generalization}
\paragraph{Across adversarial patch attacks.}
With the same defense configuration, RECLAIM reaches 53.75\% SR under EDPA~\cite{edpa}, 54.58\% under RFAS~\cite{rfas}, and 38.33\% under UPA~\cite{uada}, as summarized in Table~\ref{tab:generalization}(a). The gains over the corresponding undefended policies are 50.42, 54.58, and 29.58 pp, with ARR values of 77.07\%, 79.39\%, and 49.31\%. Long is already harder in the clean setting, with 50.00\% SR versus 75.00\% across the other three suites. Patches further reduce undefended Long SR to 0.00--3.33\%; RECLAIM restores it to 25.00--43.33\%. Across Spatial, Object, and Goal, defended SR reaches 57.78\% for UADA, 57.22\% for EDPA, 58.33\% for RFAS, and 42.78\% for UPA. Recovery thus extends across all four attacks in both the complete benchmark and the explicitly separated three-suite subset. Each comparison uses matched initial conditions and pre-generated patches.

\noindent\begin{minipage}{\columnwidth}
\centering
\captionsetup{skip=2pt}
\captionof{table}{\textbf{Generalization along two dimensions: rollout SR (\%).} All includes four suites; S/O/G excludes Long. Panel (a) uses matched conditions.}
\label{tab:generalization}
\small
\setlength{\tabcolsep}{3pt}
\begin{tabular}{lrrrr}
\toprule
\multicolumn{5}{l}{\textbf{(a) Across patch attacks: OpenVLA}} \\
\midrule
 & \multicolumn{2}{c}{All suites} & \multicolumn{2}{c}{S/O/G} \\
\cmidrule(lr){2-3}\cmidrule(lr){4-5}
Attack & Undef. & RECLAIM & Undef. & RECLAIM \\
\midrule
UADA & 0.00 & 52.92 & 0.00 & 57.78 \\
EDPA & 3.33 & 53.75 & 3.33 & 57.22 \\
RFAS & 0.00 & 54.58 & 0.00 & 58.33 \\
UPA & 8.75 & 38.33 & 11.67 & 42.78 \\
\bottomrule
\end{tabular}
\par\smallskip
\begin{tabular}{llrr}
\toprule
\multicolumn{4}{l}{\textbf{(b) Across VLA policies}} \\
\midrule
Policy & Condition & All & S/O/G \\
\midrule
OpenVLA & Clean & 68.75 & 75.00 \\
 & UADA, undef. & 0.00 & 0.00 \\
 & UADA, RECLAIM & 52.92 & 57.78 \\
\midrule
NORA & Clean & 61.25 & 80.00 \\
 & UADA, undef. & 37.08 & 49.44 \\
 & UADA, RECLAIM & 40.12 & 53.49 \\
\bottomrule
\end{tabular}
\end{minipage}\par\smallskip

\paragraph{Across VLA policies.}
We evaluate NORA~\cite{nora} under the same UADA attack setting as OpenVLA, retaining native decoding, normalization, and policy-specific calibration (Table~\ref{tab:generalization}(b)). RECLAIM achieves 53.49\% SR on S/O/G and 40.12\% across all four suites. NORA's clean Long SR is only 5.00\%, versus 80.00\% on S/O/G; the latter therefore provides a more informative assessment of defense performance within the policy's task capability. Together with the OpenVLA results, these findings support RECLAIM's ability to protect different VLA models.

\section{Conclusion}
\label{sec:conclusion}

Generalist policies increasingly span diverse tasks and embodiments~\cite{openx,octo,pi0}, making reliable visual feedback important for sustained task execution.

We presented RECLAIM, a training-free defense that uses a frozen VLA's attention and action responses to guide local visual interventions. Robot geometry contextualizes attention, while treatment comparisons and temporal revalidation support continued task execution. On four LIBERO suites, RECLAIM recovers OpenVLA performance across four patch attacks and outperforms accelerated PAD under UADA in both success rate and mean decision cost. The accompanying improvement in mask alignment indicates more accurate placement of visual interventions along the evaluated trajectories. These findings highlight how embodied attention feedback can guide visual purification as robot actions change subsequent observations, connecting adversarial patch defense to continued task execution within the perception--action loop. This motivates further study of policy feedback for robust embodied control.

\clearpage
\flushbottom
\section*{LLM Usage Statement}
We used OpenAI ChatGPT and Codex for manuscript feedback and for drafting and revising prose based on author-provided research content. These tools assisted with related-work organization, citation checks against primary sources, analysis and organization of existing experimental records and data, and refinement of figure presentation and LaTeX layout. Codex also generated and debugged implementation and experiment-support code, which the authors reviewed and revised through iterative use. Reported measurements were obtained from executed experiments and their logs. The authors retain full responsibility for the research decisions, scientific claims, references, code, and final manuscript.

{\small\bibliographystyle{ieeenat_fullname}\bibliography{references}}

@inproceedings{openvla,
  author = {Kim, Moo Jin and Pertsch, Karl and Karamcheti, Siddharth and Xiao, Ted and Balakrishna, Ashwin and Nair, Suraj and Rafailov, Rafael and Foster, Ethan P. and Sanketi, Pannag R. and Vuong, Quan and Kollar, Thomas and Burchfiel, Benjamin and Tedrake, Russ and Sadigh, Dorsa and Levine, Sergey and Liang, Percy and Finn, Chelsea},
  title = {{OpenVLA}: An Open-Source Vision-Language-Action Model},
  booktitle = {Proceedings of the 8th Conference on Robot Learning},
  series = {Proceedings of Machine Learning Research},
  volume = {270},
  pages = {2679--2713},
  year = {2025},
  publisher = {PMLR},
  url = {https://proceedings.mlr.press/v270/kim25c.html}
}

@inproceedings{libero,
  author = {Liu, Bo and Zhu, Yifeng and Gao, Chongkai and Feng, Yihao and Liu, Qiang and Zhu, Yuke and Stone, Peter},
  title = {{LIBERO}: Benchmarking Knowledge Transfer for Lifelong Robot Learning},
  booktitle = {Advances in Neural Information Processing Systems},
  volume = {36},
  year = {2023},
  doi = {10.52202/075280-1939},
  url = {https://papers.neurips.cc/paper_files/paper/2023/hash/8c3c666820ea055a77726d66fc7d447f-Abstract-Datasets_and_Benchmarks.html}
}

@inproceedings{rt2,
  author = {Zitkovich, Brianna and Yu, Tianhe and Xu, Sichun and Xu, Peng and Xiao, Ted and Xia, Fei and Wu, Jialin and Wohlhart, Paul and Welker, Stefan and Wahid, Ayzaan and Vuong, Quan and Vanhoucke, Vincent and Tran, Huong and Soricut, Radu and Singh, Anikait and Singh, Jaspiar and Sermanet, Pierre and Sanketi, Pannag R. and Salazar, Grecia and Ryoo, Michael S. and Reymann, Krista and Rao, Kanishka and Pertsch, Karl and Mordatch, Igor and Michalewski, Henryk and Lu, Yao and Levine, Sergey and Lee, Lisa and Lee, Tsang-Wei Edward and Leal, Isabel and Kuang, Yuheng and Kalashnikov, Dmitry and Julian, Ryan and Joshi, Nikhil J. and Irpan, Alex and Ichter, Brian and Hsu, Jasmine and Herzog, Alexander and Hausman, Karol and Gopalakrishnan, Keerthana and Fu, Chuyuan and Florence, Pete and Finn, Chelsea and Dubey, Kumar Avinava and Driess, Danny and Ding, Tianli and Choromanski, Krzysztof Marcin and Chen, Xi and Chebotar, Yevgen and Carbajal, Justice and Brown, Noah and Brohan, Anthony and Arenas, Montserrat Gonzalez and Han, Kehang},
  title = {{RT-2}: Vision-Language-Action Models Transfer Web Knowledge to Robotic Control},
  booktitle = {Proceedings of the 7th Conference on Robot Learning},
  series = {Proceedings of Machine Learning Research},
  volume = {229},
  pages = {2165--2183},
  year = {2023},
  publisher = {PMLR},
  url = {https://proceedings.mlr.press/v229/zitkovich23a.html}
}

@inproceedings{uada,
  author = {Wang, Taowen and Han, Cheng and Liang, James and Yang, Wenhao and Liu, Dongfang and Zhang, Luna Xinyu and Wang, Qifan and Luo, Jiebo and Tang, Ruixiang},
  title = {Exploring the Adversarial Vulnerabilities of Vision-Language-Action Models in Robotics},
  booktitle = {Proceedings of the IEEE/CVF International Conference on Computer Vision},
  pages = {6948--6958},
  year = {2025},
  url = {https://openaccess.thecvf.com/content/ICCV2025/html/Wang_Exploring_the_Adversarial_Vulnerabilities_of_Vision-Language-Action_Models_in_Robotics_ICCV_2025_paper.html}
}

@article{edpa,
  author = {Xu, Haochuan and Koh, Yun Sing and Huang, Shuhuai and Zhou, Zirun and Wang, Di and Sakuma, Jun and Zhang, Jingfeng},
  title = {Model-agnostic Adversarial Attack and Defense for Vision-Language-Action Models},
  journal = {arXiv preprint arXiv:2510.13237},
  year = {2025},
  eprint = {2510.13237},
  archivePrefix = {arXiv},
  primaryClass = {cs.CV},
  doi = {10.48550/arXiv.2510.13237},
  url = {https://arxiv.org/abs/2510.13237}
}

@inproceedings{pad,
  author = {Jing, Lihua and Wang, Rui and Ren, Wenqi and Dong, Xin and Zou, Cong},
  title = {{PAD}: Patch-Agnostic Defense against Adversarial Patch Attacks},
  booktitle = {Proceedings of the IEEE/CVF Conference on Computer Vision and Pattern Recognition},
  pages = {24472--24481},
  year = {2024},
  url = {https://openaccess.thecvf.com/content/CVPR2024/html/Jing_PAD_Patch-Agnostic_Defense_against_Adversarial_Patch_Attacks_CVPR_2024_paper.html}
}

@inproceedings{byovla,
  author = {Hancock, Asher J. and Ren, Allen Z. and Majumdar, Anirudha},
  title = {Run-time Observation Interventions Make Vision-Language-Action Models More Visually Robust},
  booktitle = {Proceedings of the IEEE International Conference on Robotics and Automation},
  pages = {9499--9506},
  year = {2025},
  doi = {10.1109/ICRA55743.2025.11128017},
  url = {https://doi.org/10.1109/ICRA55743.2025.11128017}
}

@inproceedings{localizationheads,
  author = {Kang, Seil and Kim, Jinyeong and Kim, Junhyeok and Hwang, Seong Jae},
  title = {Your Large Vision-Language Model Only Needs A Few Attention Heads For Visual Grounding},
  booktitle = {Proceedings of the IEEE/CVF Conference on Computer Vision and Pattern Recognition},
  pages = {9339--9350},
  year = {2025},
  url = {https://openaccess.thecvf.com/content/CVPR2025/html/Kang_Your_Large_Vision-Language_Model_Only_Needs_A_Few_Attention_Heads_CVPR_2025_paper.html}
}

@inproceedings{resclip,
  author = {Yang, Yuhang and Deng, Jinhong and Li, Wen and Duan, Lixin},
  title = {{ResCLIP}: Residual Attention for Training-free Dense Vision-language Inference},
  booktitle = {Proceedings of the IEEE/CVF Conference on Computer Vision and Pattern Recognition},
  pages = {29968--29978},
  year = {2025},
  url = {https://openaccess.thecvf.com/content/CVPR2025/html/Yang_ResCLIP_Residual_Attention_for_Training-free_Dense_Vision-language_Inference_CVPR_2025_paper.html}
}

@article{telea,
  author = {Telea, Alexandru},
  title = {An Image Inpainting Technique Based on the Fast Marching Method},
  journal = {Journal of Graphics Tools},
  volume = {9},
  number = {1},
  pages = {23--34},
  year = {2004},
  doi = {10.1080/10867651.2004.10487596},
  url = {https://doi.org/10.1080/10867651.2004.10487596}
}

@inproceedings{vit,
  author = {Dosovitskiy, Alexey and Beyer, Lucas and Kolesnikov, Alexander and Weissenborn, Dirk and Zhai, Xiaohua and Unterthiner, Thomas and Dehghani, Mostafa and Minderer, Matthias and Heigold, Georg and Gelly, Sylvain and Uszkoreit, Jakob and Houlsby, Neil},
  title = {An Image is Worth 16x16 Words: Transformers for Image Recognition at Scale},
  booktitle = {International Conference on Learning Representations},
  year = {2021},
  url = {https://openreview.net/forum?id=YicbFdNTTy}
}

@inproceedings{siglip,
  author = {Zhai, Xiaohua and Mustafa, Basil and Kolesnikov, Alexander and Beyer, Lucas},
  title = {Sigmoid Loss for Language Image Pre-Training},
  booktitle = {Proceedings of the IEEE/CVF International Conference on Computer Vision},
  pages = {11975--11986},
  year = {2023},
  url = {https://openaccess.thecvf.com/content/ICCV2023/html/Zhai_Sigmoid_Loss_for_Language_Image_Pre-Training_ICCV_2023_paper.html}
}

@article{dinov2,
  author = {Oquab, Maxime and Darcet, Timoth{\'e}e and Moutakanni, Th{\'e}o and Vo, Huy V. and Szafraniec, Marc and Khalidov, Vasil and Fernandez, Pierre and Haziza, Daniel and Massa, Francisco and El-Nouby, Alaaeldin and Assran, Mahmoud and Ballas, Nicolas and Galuba, Wojciech and Howes, Russell and Huang, Po-Yao and Li, Shang-Wen and Misra, Ishan and Rabbat, Michael and Sharma, Vasu and Synnaeve, Gabriel and Xu, Hu and J{\'e}gou, Herv{\'e} and Mairal, Julien and Labatut, Patrick and Joulin, Armand and Bojanowski, Piotr},
  title = {{DINOv2}: Learning Robust Visual Features without Supervision},
  journal = {Transactions on Machine Learning Research},
  year = {2024},
  url = {https://openreview.net/forum?id=a68SUt6zFt}
}

@article{shannon,
  author = {Shannon, Claude E.},
  title = {A Mathematical Theory of Communication},
  journal = {Bell System Technical Journal},
  volume = {27},
  number = {3},
  pages = {379--423},
  year = {1948},
  doi = {10.1002/j.1538-7305.1948.tb01338.x},
  url = {https://doi.org/10.1002/j.1538-7305.1948.tb01338.x}
}

@inproceedings{diffender,
  author = {Kang, Caixin and Dong, Yinpeng and Wang, Zhengyi and Ruan, Shouwei and Chen, Yubo and Su, Hang and Wei, Xingxing},
  title = {{DIFFender}: Diffusion-Based Adversarial Defense against Patch Attacks},
  booktitle = {Computer Vision -- ECCV 2024},
  pages = {130--147},
  doi = {10.1007/978-3-031-72943-0_8},
  url = {https://doi.org/10.1007/978-3-031-72943-0_8},
  year = {2024}
}

@inproceedings{apex,
  author = {Wang, Jiayimei and Ni, Tao and Xu, Guowen and Zhao, Qingchuan and Wang, Cong},
  title = {Adversarial Patch {EXterminator}: Zero-Shot and Patch-Agnostic Defense Framework Against Adversarial Patch Attacks},
  booktitle = {35th USENIX Security Symposium (USENIX Security 26)},
  year = {2026},
  pages = {4463--4482},
  publisher = {USENIX Association},
  url = {https://www.usenix.org/conference/usenixsecurity26/presentation/wang-jiayimei}
}

@article{harris2020array,
  author = {Harris, Charles R. and Millman, K. Jarrod and van der Walt, St{\'e}fan J. and Gommers, Ralf and Virtanen, Pauli and Cournapeau, David and Wieser, Eric and Taylor, Julian and Berg, Sebastian and Smith, Nathaniel J. and Kern, Robert and Picus, Matti and Hoyer, Stephan and van Kerkwijk, Marten H. and Brett, Matthew and Haldane, Allan and Fern{\'a}ndez del R{\'i}o, Jaime and Wiebe, Mark and Peterson, Pearu and G{\'e}rard-Marchant, Pierre and Sheppard, Kevin and Reddy, Tyler and Weckesser, Warren and Abbasi, Hameer and Gohlke, Christoph and Oliphant, Travis E.},
  title = {Array programming with {NumPy}},
  journal = {Nature},
  volume = {585},
  number = {7825},
  pages = {357--362},
  year = {2020},
  doi = {10.1038/s41586-020-2649-2},
  url = {https://doi.org/10.1038/s41586-020-2649-2}
}

@article{brown2017patch,
  author = {Brown, Tom B. and Man{\'e}, Dandelion and Roy, Aurko and Abadi, Mart{\'i}n and Gilmer, Justin},
  title = {Adversarial Patch},
  journal = {arXiv preprint arXiv:1712.09665},
  year = {2017},
  url = {https://arxiv.org/abs/1712.09665}
}

@inproceedings{rfas,
  author = {Lu, Hui and Yu, Yi and Yang, Yiming and Yi, Chenyu and Zhang, Qixin and Shen, Bingquan and Kot, Alex C. and Jiang, Xudong},
  title = {When Robots Obey the Patch: Universal Transferable Patch Attacks on Vision-Language-Action Models},
  booktitle = {Proceedings of the IEEE/CVF Conference on Computer Vision and Pattern Recognition},
  pages = {22867--22878},
  year = {2026},
  url = {https://openaccess.thecvf.com/content/CVPR2026/html/Lu_When_Robots_Obey_the_Patch_Universal_Transferable_Patch_Attacks_on_CVPR_2026_paper.html}
}

@inproceedings{sam,
  author = {Kirillov, Alexander and Mintun, Eric and Ravi, Nikhila and Mao, Hanzi and Rolland, Chloe and Gustafson, Laura and Xiao, Tete and Whitehead, Spencer and Berg, Alexander C. and Lo, Wan-Yen and Doll{\'a}r, Piotr},
  title = {Segment Anything},
  booktitle = {Proceedings of the IEEE/CVF International Conference on Computer Vision},
  pages = {4015--4026},
  year = {2023},
  url = {https://openaccess.thecvf.com/content/ICCV2023/html/Kirillov_Segment_Anything_ICCV_2023_paper.html}
}

@article{lin1991divergence,
  author = {Lin, Jianhua},
  title = {Divergence Measures Based on the Shannon Entropy},
  journal = {IEEE Transactions on Information Theory},
  volume = {37},
  number = {1},
  pages = {145--151},
  year = {1991},
  doi = {10.1109/18.61115},
  url = {https://doi.org/10.1109/18.61115}
}

@article{nora,
  author = {Hung, Chia-Yu and Sun, Qi and Hong, Pengfei and Zadeh, Amir and Li, Chuan and Tan, U-Xuan and Majumder, Navonil and Poria, Soujanya},
  title = {{NORA}: A Small Open-Sourced Generalist Vision Language Action Model for Embodied Tasks},
  journal = {arXiv preprint arXiv:2504.19854},
  year = {2025},
  url = {https://arxiv.org/abs/2504.19854}
}

@inproceedings{openx,
  author = {{Open X-Embodiment Collaboration}},
  title = {Open {X}-Embodiment: Robotic Learning Datasets and {RT-X} Models},
  booktitle = {Proceedings of the IEEE International Conference on Robotics and Automation},
  pages = {6892--6903},
  year = {2024},
  doi = {10.1109/ICRA57147.2024.10611477},
  url = {https://robotics-transformer-x.github.io/}
}

@inproceedings{octo,
  author = {Ghosh, Dibya and Walke, Homer Rich and Pertsch, Karl and Black, Kevin and Mees, Oier and Dasari, Sudeep and Hejna, Joey and Kreiman, Tobias and Xu, Charles and Luo, Jianlan and Tan, You Liang and Chen, Lawrence Yunliang and Vuong, Quan and Xiao, Ted and Sanketi, Pannag R and Sadigh, Dorsa and Finn, Chelsea and Levine, Sergey},
  title = {{Octo}: An Open-Source Generalist Robot Policy},
  booktitle = {Proceedings of Robotics: Science and Systems},
  year = {2024},
  doi = {10.15607/RSS.2024.XX.090},
  url = {https://roboticsproceedings.org/rss20/p090.html}
}

@inproceedings{pi0,
  author = {Black, Kevin and Brown, Noah and Driess, Danny and Esmail, Adnan and Equi, Michael Robert and Finn, Chelsea and Fusai, Niccolo and Groom, Lachy and Hausman, Karol and Ichter, Brian and Jakubczak, Szymon and Jones, Tim and Ke, Liyiming and Levine, Sergey and Li-Bell, Adrian and Mothukuri, Mohith and Nair, Suraj and Pertsch, Karl and Shi, Lucy Xiaoyang and Smith, Laura and Tanner, James and Vuong, Quan and Walling, Anna and Wang, Haohuan and Zhilinsky, Ury},
  title = {{$\pi_0$}: A Vision-Language-Action Flow Model for General Robot Control},
  booktitle = {Proceedings of Robotics: Science and Systems},
  year = {2025},
  doi = {10.15607/RSS.2025.XXI.010},
  url = {https://roboticsproceedings.org/rss21/p010.html}
}
\end{document}